\documentclass[conference]{IEEEtran} 

\usepackage{cite}
\usepackage{amsmath,amssymb,amsfonts}
\usepackage{algorithmicx}
\usepackage[pdftex]{graphicx} 
\usepackage{textcomp} 
\usepackage{xcolor}
\def\BibTeX{{\rm B\kern-.05em{\sc i\kern-.025em b}\kern-.08em
    T\kern-.1667em\lower.7ex\hbox{E}\kern-.125emX}}

\usepackage{hyperref}

\usepackage{cleveref}
\Crefname{equation}{Eq.}{Eqs.} 
\Crefname{figure}{Fig.}{Figs.}

\graphicspath{{images-src/}{images-bin/}} 

\definecolor{dlrprim1}{HTML}{000000} 
\definecolor{dlrprim2}{HTML}{666666} 
\definecolor{dlrprim3}{HTML}{b9cad2}
\definecolor{dlrprim4}{HTML}{ffffff} 

\definecolor{dlrblue1}{HTML}{00658b} 
\definecolor{dlrblue2}{HTML}{3b98cb}
\definecolor{dlrblue3}{HTML}{6cb9dc}
\definecolor{dlrblue4}{HTML}{a7d3ec}
\definecolor{dlrblue5}{HTML}{d1e8fa}

\definecolor{dlryellow1}{HTML}{d2ae3d}  
\definecolor{dlryellow2}{HTML}{f2cd51} 
\definecolor{dlryellow3}{HTML}{f8de53}
\definecolor{dlryellow3}{HTML}{fcea7a}
\definecolor{dlryellow3}{HTML}{fff8be}

\definecolor{dlrgreen1}{HTML}{82a043} 
\definecolor{dlrgreen2}{HTML}{a6bf51}
\definecolor{dlrgreen3}{HTML}{cad55c}
\definecolor{dlrgreen4}{HTML}{d9df78}
\definecolor{dlrgreen5}{HTML}{e6eaaf}

\definecolor{dlrgray1}{HTML}{666666} 
\definecolor{dlrgray2}{HTML}{868585}
\definecolor{dlrgray3}{HTML}{b1b1b1}
\definecolor{dlrgray4}{HTML}{cfcfcf}
\definecolor{dlrgray5}{HTML}{ebebeb}

\usepackage[textsize=scriptsize]{todonotes} 
\newcommand{\annotategraphicsmulti}[3][]{
  \begin{tikzpicture}[%
  every node/.style={draw=black, black, text opacity=1, fill=white, fill opacity=0.75,inner sep=0.5mm, #1},%
  ]
  \node[anchor=south west,inner sep=0, draw=none] (image) at (0,0) {
    #2
  };
  \begin{scope}[x={(image.south east)},y={(image.north west)}]
    #3
  \end{scope}
  \end{tikzpicture}%
}

\newcommand*\lref[1]{\tikz[baseline=(char.base)]{\node[shape=rectangle,draw,inner sep=2pt] (char) {\scriptsize #1};}}

\usepackage{subcaption}
\usepackage{bm}
\newcommand{\Log}{\mathrm{Log}}
\newcommand{\Exp}{\mathrm{Exp}}
\newcommand{\SE}{\mathrm{SE}(3)}

\usepackage{siunitx}
\DeclareSIUnit\year{yrs.}

\usepackage{algorithm}
\usepackage{algpseudocode}
\algdef{SE}[DOWHILE]{Do}{doWhile}{\algorithmicdo}[1]{\algorithmicwhile\ #1}%

\usepackage[acronym]{glossaries}
\glsdisablehyper
\newacronym{sts}{STS}{single tangent space}
\newacronym{gmm}{GMM}{Gaussian Mixture Model}
\newacronym{gmr}{GMR}{Gaussian Mixture Regression}
\newacronym{lqt}{LQT}{Linear Quadratic Tracking}
\newacronym{ilqr}{iLQR}{iterative Linear Quadratic Regulation}
\newacronym[longplural=Gaussian Processes]{gp}{GP}{Gaussian Process}
\newacronym{kmp}{KMP}{Kernelized Movement Primitive}
\newacronym{rbf}{RBF}{Radial Basis Function}
\newacronym[longplural=degrees of freedom]{dof}{DoF}{degree of freedom}
\newacronym{ml}{ML}{Machine Learning}
\newacronym{lfd}{LfD}{learning from demonstration}
\newacronym{ds}{DS}{Dynamical System}
\newacronym{poe}{PoE}{product of experts}
\newacronym{vf}{VF}{Virtual Fixture}

\newacronym{dvrk}{dVRK}{da Vinci Research Kit}
\newacronym{mtm}{MTM}{Master Tool Manipulator}
\newacronym{psm}{PSM}{Patient Side Manipulator}

\newacronym{emg}{EMG}{Electromyography}
\newacronym{mvc}{MVC}{maximum voluntary contraction}

\newcommand{\tumaffiliation}{Department of Computer Engineering, Technical University of Munich, Friedrich-Ludwig-Bauer-Str. 3, Garching, Germany.}
\newcommand{\rmcaffiliation}{German Aerospace Center (DLR), Robotics and Mechatronics Center (RMC), M\"unchener Str. 20, 82234 We\ss ling, Germany.}
\newcommand{\napoliaffiliation}{Department of Electrical Engineering and Information Technology (DIETI), University of Naples Federico II, 80131 Naples, Italy}
\newcommand{\createaffiliation}{C.R.E.A.T.E. Consorzio di Ricerca per l'Energia, l'Automazione e le Tecnologie dell'Elettromagnetismo, Università degli Studi di Napoli Federico II, Napoli, Italy.}
\newcommand{\genoaaffiliation}{Department of Mechanical, Energy, Management and Transportation Engineering (DIME), University of Genova, Genoa, Italy}
\newcommand{\idiapaffiliation}{Idiap Research Institute, Martigny, Switzerland.}
\newcommand{\epflaffiliation}{École Polytechnique Fédérale de Lausanne (EPFL), Switzerland.}
\newcommand{\lisbonaffiliation}{Instituto Superior Técnico, University of Lisbon, Portugal.}

\IEEEoverridecommandlockouts                             

\title{Learning Geometry-Aware Virtual Fixtures From Sparse Demonstrations}

\author{Maximilian Mühlbauer$^{1,2}$, Marcella Piacentino$^3$, Raffaella Mancino$^4$, Paolino De Risi$^5$, Bernhard Weber$^1$,\\ Margarida Campos$^{1,6}$, Thomas Hulin$^1$, Sylvain Calinon$^{7,8}$, Freek Stulp$^1$, Alin Albu-Schäffer$^{1,2}$,\\ Julian Klodmann$^1$, Fanny Ficuciello$^5$, João Silvério$^1$
\thanks{$^1$ \rmcaffiliation}
\thanks{$^2$ \tumaffiliation}
\thanks{$^3$ \createaffiliation}
\thanks{$^4$ \genoaaffiliation}
\thanks{$^5$ \napoliaffiliation}
\thanks{$^6$ \lisbonaffiliation}
\thanks{$^7$ \idiapaffiliation}
\thanks{$^8$ \epflaffiliation}
}

\begin{document}

\maketitle


\begin{abstract}
In many teleoperation applications, collecting a large number of demonstrations as required for traditional probabilistic \gls{lfd} approaches may not be feasible.
To still give operators the ability to intuitively create trajectories as \glspl{vf}, we propose to leverage a \textit{motion prior} in the learning process.
Particularly, by using \gls{lqt}, users are able to define guiding trajectories from the demonstration of just a few via points.
To account for orientation guidance, we further reformulate classical \gls{lqt} on Riemannian manifolds, introducing an additional \textit{geometric} prior.
Through a probabilistic interpretation of the \gls{lqt} solution, we derive a covariance estimate at each trajectory point which we use to modulate the stiffness of the resulting fixture, resulting in \textit{strong} guidance around the via points and \textit{softer} guidance when far away.
The covariance information is also used to define a validity region of the fixture, allowing the operator to leave its influence area and conduct unmodeled tasks.
We evaluate the proposed Riemannian \gls{lqt} formulation in a set of toy examples and the full framework on a cutting task requiring high precision in a minimally invasive surgery setting on the \textit{\acrlong{dvrk}} (\acrshort{dvrk}).
\end{abstract}

\begin{IEEEkeywords}
Physical Human-Robot Interaction, Medical Robots and Systems, Learning from Demonstration
\end{IEEEkeywords}

\section{Introduction}
\begin{figure}
	\centering
	\includegraphics[width=\columnwidth,page=5,trim={0 330 0 0},clip]{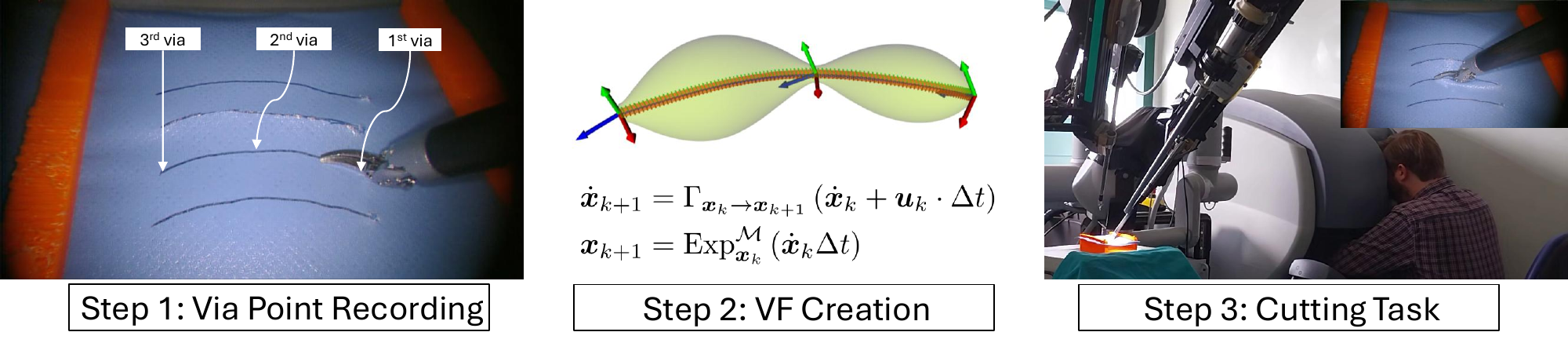}
	\caption{\label{fig:intro:overview} Overview of our approach. Users record probabilistic \textit{via points} which are then used to fit a \textit{Virtual Fixture} with a linear dynamics motion prior. The fixture is then used to guide a user during a \textit{cutting task}.\vspace{-1em}}
\end{figure}

\glsresetall
\glspl{vf} \cite{rosenberg1993virtualfixtures,bowyer2014active} provide augmented force feedback to a human operator in hands-on or teleoperated manipulation settings and are especially important when high precision has to be achieved to successfully complete a task.
Even when the fixtures themselves are slightly imprecise, users only have to correct misaligned \glspl{dof} while guidance along the remaining \glspl{dof} remains available~\cite{muehlbauer2022multiphase}.
One popular approach in recent years to intuitively create \glspl{vf} is by using \gls{lfd} \cite{raiola2017comanipulation,zeestraten2018programming,muehlbauer2024probabilistic}.
These approaches do, however, require multiple demonstrations to create fixtures which is challenging when guidance should be available for unique tasks or when the operation to be performed, such as cutting, is destructive and thus cannot be repeated.
\Cref{fig:intro:overview} shows such use case, where the task is to cut a tissue phantom in a minimally invasive surgical setting using the \gls{dvrk}~\cite{kazanzides2014davinci}.

To solve this task, namely, to provide \gls{vf} guidance to the human operator from demonstrating only few via points, we leverage a \textit{motion prior}.
To this end, we employ \gls{lqt} which assumes a double-integrator dynamical system passing through via points with a specified accuracy and a cost on control commands, thus favoring smooth velocity changes.
Traditionally, however, \gls{lqt} has been formulated for linear systems on Euclidean data, which does not allow for the inclusion of orientation data.
We therefore first derive a formulation of \gls{lqt} on Riemannian manifolds in \Cref{sec:method:lqt}, thus introducing a \textit{geometric} prior.

The trajectory resulting from state evolution can then be used as \gls{vf} as depicted in \Cref{fig:intro:overview} and shown in \Cref{sec:method:fixture}.
Using a teleoperation system (\Cref{sec:method:teleop}), an operator commands the surgical robots from haptic input devices.
A fixture in remote robot coordinates rendered locally provides both positional and orientational guidance with variable stiffness to the operator (\Cref{sec:method:control}).
We evaluate both fundamental properties of the proposed formulation and the real-robot implementation of the whole system on the \gls{dvrk} using a pilot study with $10$ participants in \Cref{sec:evaluation}.

In summary, our contributions are the following:
\begin{enumerate}
	\item A formulation of \gls{lqt} using the Riemannian Gaussian distribution~\cite{zeestraten2017manifold,calinon2020gaussians} on Riemannian manifolds to incorporate orientation,
	\item the creation of \Acrlongpl{vf} using only few demonstrated points with a strong \textit{motion prior} using \gls{lqt},
	\item a teleoperation system for rendering \glspl{vf} defined in remote robot coordinates locally as well as their passivation.
\end{enumerate}
We provide background on the methods in \Cref{sec:background}.

\section{Related Work}
While defining \glspl{vf} based on geometric primitives and rendering them to the user using impedance control is a mature technology \cite{bowyer2014active}, the adaptive and interactive creation of constraints is still an open research topic.
In \cite{pruks2022method}, vision is used to detect 2D features; users can then place fixtures based on primitive shapes onto the detected location.
Visual detection of the actual target objects is used by \cite{selvaggio2016enhancing} to dynamically define \glspl{vf}.
Both approaches are, however, not designed for continuous interaction as required in cutting tasks but rather provide constraints to reach an object to be manipulated.

An overview of recent \gls{vf} methods used in minimally invasive surgery is provided by \cite{menoth2025review}.
Closest to our approach are \cite{selvaggio2018passive} and \cite{moccia2019vision}, where in the former users teach a fixture based on via points used as control points for a spline and the latter extends it via an automatic extraction of via points from image data, both applying it to a cutting task.
Both methods -- similar to our motion prior -- create smooth fixtures, however, without
\textit{uncertainty-aware} guidance and only assisting in the $xy$-plane whereas we aim to provide a full $6$ \gls{dof} guidance.

To incorporate \textit{probabilistic} guidance, as opposed to a simple spline interpolation, we leverage \gls{lqt} which tracks a set of desired points with minimal intervention control \cite{calinon2015tutorial}, providing both mean and covariance for resulting states.
Originally, the formulation only allows for Euclidean states and velocities.
A variant of \gls{lqt} on a single tangent space, linearizing the manifold around the initial state, has been proposed by \cite{calinon2020gaussians}, which, however, suffers from imprecisions as we show in \Cref{sec:method:lqt_eval}.
Closest to our \gls{lqt} formulation is \cite{rozo2020learning}, where the authors use a different cost function for via point tracking; they furthermore do not derive the optimization procedure or give a probabilistic interpretation of the result.

\section{Background}
\label{sec:background}
\subsection[The da Vinci Research Kit]{The \Acrlong{dvrk}}
\begin{figure}
	\centering
	\includegraphics[width=\columnwidth,page=2,trim={140 35 130 65},clip]{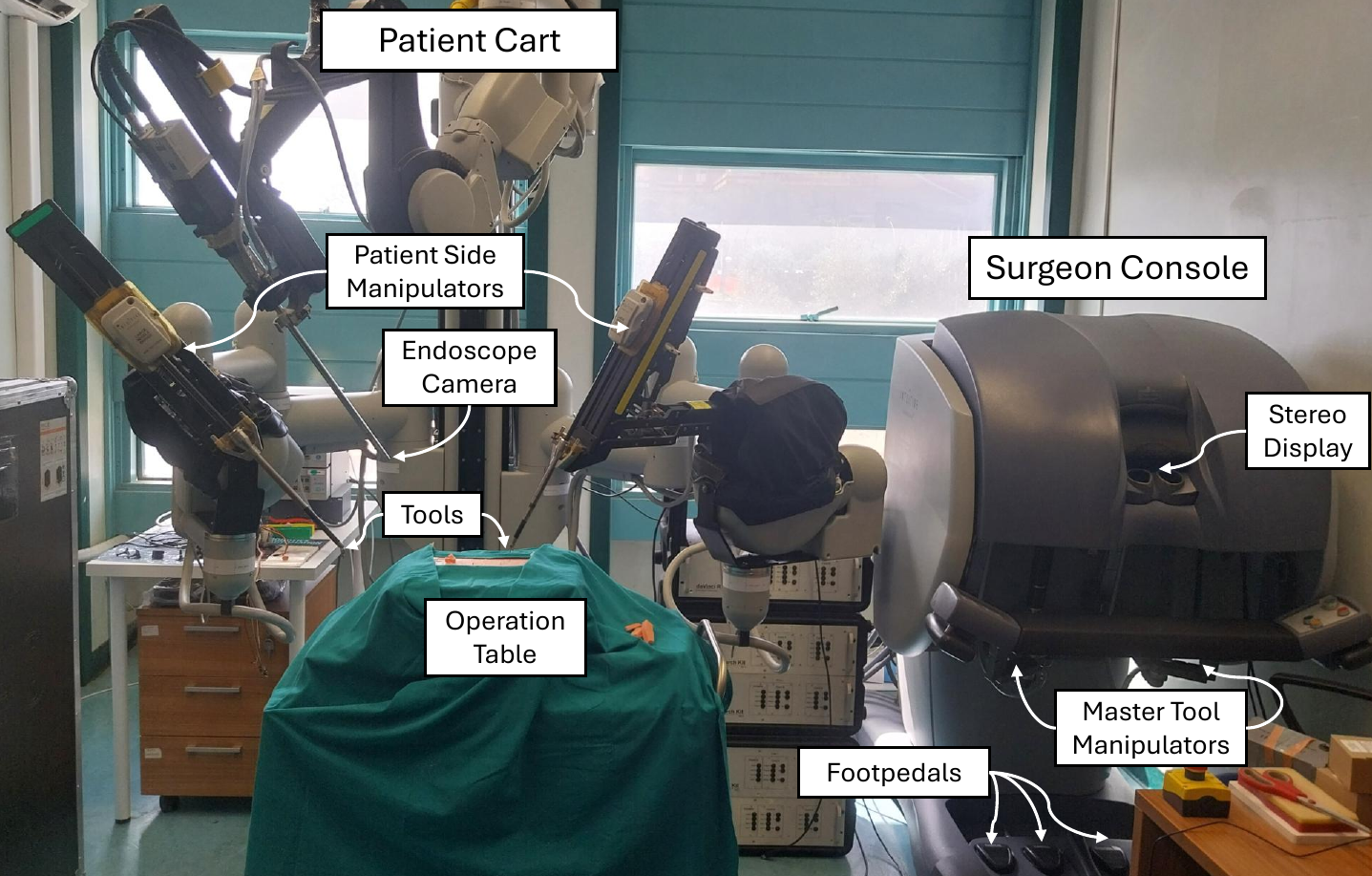}
	\caption{\label{fig:background:davinci} The \Acrlong{dvrk} used in the experiments.\vspace{-1em}}
\end{figure}
An overview of the \gls{dvrk} \cite{kazanzides2014davinci,ferro2023coppelia}, a research robot derived from the da Vinci minimally invasive surgery system which we use in the evaluation (\Cref{sec:evaluation}), is shown in \Cref{fig:background:davinci}.
We use the teleoperation provided by the system, consisting of position-controlled \glspl{psm} coupled to the \glspl{mtm} in zero gravity control with aligned rotations and scaled translations.
Our \glspl{vf} are implemented as assistive forces on the \glspl{mtm}, thus constraining the \glspl{psm} through the teleoperated coupling.

\subsection{Riemannian Manifolds}
Many quantities in robotics, such as orientations in $\mathrm{SO}(3)$, cannot be modeled using Euclidean geometry.
Riemannian manifolds allow for a principled treatment of such non-Euclidean geometries \cite{zeestraten2017manifold,calinon2020gaussians}.
Product manifolds can be constructed to represent a full pose - in this work, we use the product manifold $\mathbb{R}^3 \times \mathrm{SO}(3)$.
Common to all smooth manifolds $\mathcal{M}$ is that a logarithm function $\bm{u}_{12} = \Log_{\bm{x}_1}^\mathcal{M} \left( \bm{x}_2 \right)$ maps a point $\bm{x}_2 \in \mathcal{M}$ onto the tangent space $\mathcal{T}_{\bm {x}_1}\mathcal{M}$ at $\bm{x}_1 \in \mathcal{M}$, with the exponential function locally performing the inverse, calculating $\bm{x}_{2} = \Exp_{\bm{x}_1}^\mathcal{M} \left( \bm{u}_{12} \right)$.
\textit{Parallel transport} moves tangent vectors between tangent spaces at different points on one manifold, while the Jacobian $\bm{J}_\mathcal{M}$ as given in~\cite{muehlbauer2025unified} transforms tangent vectors between tangent spaces of different manifolds $\mathcal{M}$ that are related via a smooth map $\mathcal{F}: \mathcal{M} \rightarrow \mathcal{M}_1$.
Optimization on manifolds can be performed using geodesic regression~\cite{fletcher2012geodesic} (including the multivariate case:~\cite{kim2014multvariate}), making use of \textit{parallel transport} to approximate a small update in the vicinity of the current linearization point.

The Riemannian Gaussian distribution \cite{zeestraten2017manifold,calinon2020gaussians} uses manifold-specific distances of the log map and is given as
\begin{equation}
	\mathcal{N}\left(\bm{x}\vert\bm{\mu},\bm{\Sigma}\right) = \frac{1}{\sqrt{\left(2\pi\right)^D\mathrm{det}(\bm{\Sigma})}}e^{-\frac{1}{2}\mathrm{Log}_{\bm{\mu}}^{\mathcal{M}}\!\left(\bm{x}\right)^\top\bm{\Sigma}^{-1}\mathrm{Log}_{\bm{\mu}}^{\mathcal{M}}\!\left(\bm{x}\right)},
\end{equation}
with mean $\bm{\mu}\in\mathcal{M}$, covariance $\bm{\Sigma} \in \mathcal{T}_{\bm{\mu}} \mathcal{M} \otimes \mathcal{T}_{\bm{\mu}} \mathcal{M}$ and $D$ the dimension of $\mathcal{T}_{\bm{\mu}} \mathcal{M}$.
Using the default metric of each $\mathcal{M}$ through its log map, we obtain different distances and thus different covariances depending on the chosen manifold \cite{ti2023geometric}, as similarly explored for the manipulability metric \cite{lachner2020influence}.

\subsection[Impedance-controlled Probabilistic Virtual Fixtures]{Impedance-controlled Probabilistic \Acrlongpl{vf}}
For torque-controlled manipulators such as the \glspl{mtm} of the \gls{dvrk}, Cartesian wrenches $\bm{w}$ are rendered through $\bm{\tau} = \bm{J}^\top \bm{w}$ \cite{hogan1984impedance}, where the Cartesian wrench itself evaluates to \cite{muehlbauer2024probabilistic,muehlbauer2025unified}
\begin{equation}
	\bm{w} = \bm{K}_{\mathrm{p}} \Delta \bm{x} + \bm{K}_{\mathrm{d}} \Delta \dot{\bm{x}}.\label{eq:impedance_control}
\end{equation}
The impedance characteristic is thus controlled through the stiffness $\bm{K}_{\mathrm{p}}$ and damping $\bm{K}_{\mathrm{d}}$ gains.
We use the method described in \cite{muehlbauer2025unified} to modulate the stiffness based on the inverse of the covariance of the attractor $\bm{\Sigma}^{-1}$ and the passivation approach of \cite{muehlbauer2026stabilizingarxiv} to stabilize varying stiffness gains.
Damping gains $\bm{K}_{\mathrm{d}}$ are calculated using double diagonalization \cite{albuschaeffer2003cartesianimpedance}.

\subsection[Linear Quadratic Tracking]{\Acrlong{lqt}}
\gls{lqt}, as presented in \cite{calinon2015tutorial}, tracks poses $\hat{\bm{\mu}}_k$ with covariance $\hat{\bm{\Sigma}}_k$ at time steps $k$ for a discrete double-integrator dynamical system using minimal intervention control.
With acceleration commands $\bm{u}_k$ in Euclidean space, we obtain
\begin{equation}
	\underbrace{\begin{bmatrix}
		\bm{x}_{k+1}\\
		\dot{\bm{x}}_{k+1}
	\end{bmatrix}}_{\bm{\xi}_{k+1}} =
	\underbrace{\begin{bmatrix}
		\bm{I} & \bm{I} \Delta t\\
		\bm{0} & \bm{I}
	\end{bmatrix}}_{\bm{A}}
	\underbrace{\begin{bmatrix}
		\bm{x}_{k}\\
		\dot{\bm{x}}_{k}
	\end{bmatrix}}_{\bm{\xi}_{k}} +
	\underbrace{\begin{bmatrix}
		\bm{0}\\
		\bm{I}\Delta t
	\end{bmatrix}}_{\bm{B}} \bm{u}_k.
\end{equation}
For tracking target poses, we minimize the cost function
\begin{align}
	J(\bm{u}) &= \left( \hat{\bm{\xi}}_K - \bm{\xi}_K \right)^\top \bm{\Sigma}_K^{-1} \left( \hat{\bm{\xi}}_K - \bm{\xi}_K \right)\nonumber\\
	&~~~~+ \sum_{k=1}^{K-1} \left( \left( \hat{\bm{\xi}}_k - \bm{\xi}_k \right)^\top \bm{\Sigma}_k^{-1} \left( \hat{\bm{\xi}}_k - \bm{\xi}_k \right) + \bm{u}_k^\top \bm{R}_k \bm{u}_k \right)\nonumber\\
	&= \left( \bm{\mu} - \bm{\zeta} \right)^\top \bm{\Sigma}^{-1} \left( \bm{\mu} - \bm{\zeta} \right) + \bm{U}^\top \bm{R} \bm{U}\label{eq:lqt:costfcn}
\end{align}
where stacked $\bm{\mu} = \left[ \bm{\mu}_1 \hdots \bm{\mu}_K \right]^\top$, $\bm{\Sigma} = \mathrm{blockdiag} \left( \bm{\Sigma}_1 \hdots \bm{\Sigma}_K \right)$, 
$\bm{R} = \mathrm{blockdiag} \left( \bm{R}_1 \hdots \bm{R}_{K-1} \right)$ and $\bm{\zeta}$, $\bm{U}$ from the unrolled system
\begin{align}
{\scriptsize
	\underbrace{\begin{bmatrix}
		\bm{\xi}_1\\
		\bm{\xi}_2\\
		\bm{\xi}_3\\
		\vdots\\
		\bm{\xi}_K
	\end{bmatrix}}_{\bm{\zeta}} = 
	\underbrace{\begin{bmatrix}
		\bm{I}\\
		\bm{A}\\
		\bm{A}^2\\
		\vdots\\
		\bm{A}^{K-1}
	\end{bmatrix}}_{\bm{S}^{\bm{\xi}}} \bm{\xi}_1 +
	\underbrace{\begin{bmatrix}
		\bm{0} & \bm{0} & \hdots & \bm{0}\\
		\bm{B} & \bm{0} & \hdots & \bm{0}\\
		\bm{AB} & \bm{B} & \hdots & \bm{0}\\
		\vdots & \vdots & \ddots & \vdots\\
		\bm{A}^{K-2} \bm{B} & \bm{A}^{K-3} \bm{B} & \hdots & \bm{B}
	\end{bmatrix}}_{\bm{S}^{\bm{u}}}
	\underbrace{\begin{bmatrix}
		\bm{u}_1\\
		\bm{u}_2\\
		\vdots\\
		\bm{u}_{K-1}
	\end{bmatrix}}_{\bm{U}}
}
\nonumber
\end{align}
are used, where $\bm{S}^{\bm{\xi}}$ unrolls the state evolution and $\bm{S}^{\bm{u}}$ control  actions.
Inserting into \eqref{eq:lqt:costfcn} results in
\begin{align}
	J(\bm{u}) &= \left( \bm{\mu} - \bm{S}^{\bm{\xi}} \bm{\xi}_1 - \bm{S}^{\bm{u}} \bm{U} \right)^\top \bm{\Sigma}^{-1} \left( \bm{\mu} - \bm{S}^{\bm{\xi}} \bm{\xi}_1 - \bm{S}^{\bm{u}} \bm{U} \right)\nonumber\\
	&~~~~+ \bm{U}^\top \bm{R} \bm{U}.
\end{align}
Differentiating w.r.t. $\bm{U}$ and equating the result to zero gives the optimal control inputs $\hat{\bm{U}} = \left[ \hat{\bm{u}}_1 \hdots \hat{\bm{u}}_{K-1} \right]^\top$
\begin{equation}
	\hat{\bm{U}} = \left( \bm{S}^{\bm{u}\top} \bm{\Sigma}^{-1} \bm{S}^{\bm{u}} + \bm{R} \right)^{-1} \bm{S}^{\bm{u}\top} \bm{\Sigma}^{-1} \left( \bm{\mu} - \bm{S}^{\bm{\xi}} \bm{\xi}_1 \right)\label{eq:lqt_eucl_solution}
\end{equation}
with covariance of the optimal control inputs $\hat{\bm{U}}$
\begin{equation}
	\hat{\bm{\Sigma}}^u = \left( \bm{S}^{\bm{u}\top} \bm{\Sigma}^{-1} \bm{S}^{\bm{u}} + \bm{R} \right)^{-1}.
\end{equation}
A transformation of $\hat{\bm{\Sigma}}^u$ into state space yields
\begin{equation}
	\hat{\bm{\Sigma}} = \bm{S}^{\bm{u}} \left( \bm{S}^{\bm{u}\top} \bm{\Sigma}^{-1} \bm{S}^{\bm{u}} + \bm{R} \right)^{-1} \bm{S}^{\bm{u}\top}
\end{equation}
which associates an uncertainty $\bm{\Sigma}_k$ with each state $\bm{\xi}_k$.

\section[Riemannian Linear Quadratic Tracking]{Riemannian \Acrlong{lqt}}
\label{sec:method:lqt}
On Riemannian manifolds, the state $\bm{\xi}_k$ consists of a pose $\bm{x}_k \in \mathcal{M}$ and corresponding velocity $\dot{\bm{x}}_k$ in $\mathcal{T}_{\bm{x}_k} \mathcal{M}$.
The state evolution is thus not a linear mapping but requires the evaluation of the nonlinear $\Exp$ map and the parallel transport $\Gamma_{\bm{x}_k \rightarrow \bm{x}_{k+1}}$ with respect to the Levi-Civita connection from $\mathcal{T}_{\bm{x}_k}\mathcal{M}$ to $\mathcal{T}_{\bm{x}_{k+1}}\mathcal{M}$ along the geodesic from $\bm{x}_{k}$ to $\bm{x}_{k+1}$
\begin{align}
	\dot{\bm{x}}_{k+1} &= \Gamma_{\bm{x}_k \rightarrow \bm{x}_{k+1}} \left( \dot{\bm{x}}_{k} + \bm{u}_{k} \cdot \Delta t \right)\label{eq:riem_lqt_vel_update}\\
	\bm{x}_{k+1} &= \mathrm{Exp}_{\bm{x}_{k}}^\mathcal{M} \left( \dot{\bm{x}}_{k} \Delta t \right)\label{eq:riem_lqt_pos_update}
\end{align}
where $\Delta t$ is the time between two consecutive steps and $\bm{u}_k$ the acceleration command at $k$.
In order to minimize the control actions for the given tracking task,\footnote{We only consider \textit{pose} tracking in this work - to also track \textit{velocities}, they have to be added to the cost function and its derivative.} we consider the cost function for $J \le K-2$ via points and $K - 1$ commands
\begin{equation}
	J(\bm{u}) = \sum_{j=1}^J \mathrm{Log}_{\hat{\bm{\mu}}_j}^\mathcal{M} \left( \bm{x}_{*(j)} \right)^\top \hat{\bm{\Sigma}}_j^{-1} \mathrm{Log}_{\hat{\bm{\mu}}_j}^\mathcal{M} \left( \bm{x}_{*(j)} \right) + \sum_{k=1}^{K-1} \bm{u}_k^\top \bm{R} \bm{u}_k \label{eq:lqt_riemann:costfcn}
\end{equation}
where $\bm{x}_{*(j)}$ selects the state corresponding to the $j$-th via point.
With $\bm{u}_k^* = \bm{u}_k + \bm{\epsilon}_k$, we obtain an \gls{ilqr} problem \cite{lembono2021probabilistic} with Taylor expansion
\begin{align}
	J(\bm{u} + \bm{\epsilon}) &= \sum_{j=1}^J \mathrm{Log}_{\hat{\bm{\mu}}_j, \bm{\epsilon}_{*(j)}}^\mathcal{M} \left( \bm{x}_{*(j)} \right)^\top \hat{\bm{\Sigma}}_j^{-1} \mathrm{Log}_{\hat{\bm{\mu}}_j, \bm{\epsilon}_{*(j)}}^\mathcal{M} \left( \bm{x}_{*(j)} \right)\nonumber\\
	&~~~~~~~~~~+ \sum_{k=1}^{K-1} \left(\bm{u}_k + \bm{\epsilon}_k\right)^\top \bm{R} \left(\bm{u}_k + \bm{\epsilon}_k\right) \label{eq:lqt_riemann:costfcn_taylor}
\end{align}
where we consider states modified by the update $\bm{\epsilon}_k$ to the acceleration command $\bm{u}_k$ in the $\mathrm{Log}$ function
\begin{align}
	 &\mathrm{Log}_{\hat{\bm{\mu}}_j, \bm{\epsilon}_{*(j)}}^\mathcal{M} \left( \bm{x}_{*(j)} \right) = \mathrm{Log}_{\hat{\bm{\mu}}_j}^\mathcal{M} \left( \bm{x}_{*(j)} \right)\\
	 &~~~+ \sum_{i=1}^{*(j)-2} \Gamma_{\bm{x}_{*(j)} \rightarrow \bm{\mu}_j} \Gamma_{\bm{x}_i \rightarrow \bm{x}_{*(j)}} \bm{\epsilon}_i \Delta t^2 \left( *(j) - i - 1 \right).\nonumber
\end{align}
$\Gamma_{\bm{x}_i \rightarrow \bm{x}_{*(j)}}$ again denotes the parallel transport with respect to the Levi-Civita connection, which we evaluate along the geodesics connecting system states between $\mathcal{T}_{\bm{x}_i}\mathcal{M}$ and $\mathcal{T}_{\bm{x}_{*(j)}}\mathcal{M}$.
Specifically, following \eqref{eq:riem_lqt_vel_update} and \eqref{eq:riem_lqt_pos_update}, we double-integrate the update $\bm{\epsilon}_i$ using $\Delta t$ to obtain a \textit{pose increment}.
The parallel transport from $\mathcal{T}_{\bm{x}_i}\mathcal{M}$ to $\mathcal{T}_{\bm{x}_{i+1}}\mathcal{M}$ is required to move the \textit{velocity} vector to the tangent at which it is integrated to a \textit{pose}.
As the parallel transport can be expressed by a left-multiplication of a rotation matrix for the manifolds we consider \cite{zeestraten2017manifold}, similarly to the Euclidean $\bm{S}^{\bm{u}}$, we can unroll control actions on Riemannian manifolds using
\begin{align}
\tiny
	\bm{\mathcal{S}}^{\bm{u}} = \begin{bmatrix}
		\bm{0} & \bm{0} & \hdots & \bm{0} & \bm{0}\\
		\bm{0} & \bm{0} & \hdots & \bm{0} & \bm{0}\\
		\bm{\Gamma}_{1,3} \Delta t^2 & \bm{0} & \hdots & \bm{0} & \bm{0}\\
		2 \bm{\Gamma}_{1,4} \Delta t^2 & \bm{\Gamma}_{2,4} \Delta t^2 & \hdots & \bm{0} & \bm{0}\\
		\vdots & \vdots & \ddots & \vdots & \vdots\\
		(K-2) \bm{\Gamma}_{1,K} \Delta t^2 & (K-3)\bm{\Gamma}_{2,K} \Delta t^2 & \hdots & \bm{\Gamma}_{K-2,K} \Delta t^2 & \bm{0}
	\end{bmatrix}.\small\nonumber
\end{align}
Inspired by \cite{kim2014multvariate}, we approximate the derivative of the Riemannian $\mathrm{Exp}$ by another parallel transport.
Subsequent parallel transports move the increment to the tangent space $\mathcal{T}_{\bm{\mu}_j} \mathcal{M}$ for error calculation, finally, we sum the changes from increments at all time steps.
This leads to a quadratic problem in $\bm{\epsilon}$ \eqref{eq:lqt_riemann:costfcn_taylor}.
Calculating its derivative with respect to $\bm{\epsilon}$ and setting to zero to obtain the extremum of this function gives
\vspace{-0.5em}
\begin{align}
	\bm{\epsilon} &= -\overbrace{\left( \bm{\mathcal{S}}'^{\bm{u}\top} \bm{\Gamma}_{\bm{x} \rightarrow \bm{\mu}}^\top \bm{\Sigma}^{-1} \bm{\Gamma}_{\bm{x} \rightarrow \bm{\mu}} \bm{\mathcal{S}}'^{\bm{u}} + \bm{R} \right)^{-1}}^{\bm{\Sigma}_{\bm{u}}} \nonumber\\
	&~~~~~~~~~~~\left(\bm{\mathcal{S}}'^{\bm{u}\top} \bm{\Gamma}_{\bm{x} \rightarrow \bm{\mu}}^\top \bm{\Sigma}^{-1} \mathrm{Log}_{\hat{\bm{\mu}}}^\mathcal{M} \left( \bm{x} \right) + \bm{R} \bm{u} \right),\label{eq:lqt_eps}
\end{align}
where we have removed rows in $\bm{\mathcal{S}}^{\bm{u}}$ for which no via points are available to obtain the wide matrix $\bm{\mathcal{S}}'^{\bm{u}}$ and
\begin{align}
	\bm{\Gamma}_{\bm{x} \rightarrow \bm{\mu}} &= \mathrm{blockdiag} \left( \bm{\Gamma}_{\bm{x}_{*(1)} \rightarrow \bm{\mu}_1}, \bm{\Gamma}_{\bm{x}_{*(2)} \rightarrow \bm{\mu}_2}, \hdots, \bm{\Gamma}_{\bm{x}_{*(J)} \rightarrow \bm{\mu}_J} \right),\nonumber\\
	\bm{\Sigma}^{-1} &= \mathrm{blockdiag} \left( \bm{\Sigma}_1^{-1}, \bm{\Sigma}_2^{-1}, \hdots, \bm{\Sigma}_J^{-1} \right)
\end{align}
for $J$ via points.
Iterating with $\bm{u} \leftarrow \bm{u} + \epsilon$, the optimization converges after a few steps.
To aid convergence for long trajectories on curved manifolds like $\mathcal{S}^2$ where the parallel transport $\bm{\Gamma}$ changes vastly depending on the current state estimate $\bm{x}$, we use line search to ensure decreasing cost for every optimization step.
Subsequently, the full covariance in tangent space of the mean estimate evaluates to
\begin{equation}
	\bm{\Sigma} = \bm{\mathcal{S}}^{\bm{u}} \underbrace{\left(\bm{\mathcal{S}}'^{\bm{u}\top} \bm{\Gamma}_{\bm{x} \rightarrow \bm{\mu}}^\top \bm{\Sigma}^{-1} \bm{\Gamma}_{\bm{x} \rightarrow \bm{\mu}} \bm{\mathcal{S}}'^{\bm{u}} + \bm{R} \right)^{-1}}_{\bm{\Sigma}_{\bm{u}}} \bm{\mathcal{S}}^{\bm{u}\top},\label{eq:lqt:cov}
\end{equation}
the estimate for an individual state can be extracted using $\left( \bm{\Sigma}_1, \bm{\Sigma}_2, \hdots, \bm{\Sigma}_K \right) = \mathrm{blockdiag}^{-1} \left( \bm{\Sigma} \right)$.%
\footnote{The covariance \eqref{eq:lqt:cov} of the Riemannian Gaussian \cite{zeestraten2017manifold,calinon2020gaussians} denotes how far one can move from the mean while keeping the cost function \eqref{eq:lqt_riemann:costfcn} low \cite{lembono2021probabilistic}.
As the control command of the first state only starts affecting the third state onwards, no meaningful covariance is obtained for the first two states.}

\begin{algorithm}[t]
\caption{\label{algo:lqt:fitting} Fitting a Cholesky-factorized Riemannian \gls{lqt}.}
\begin{algorithmic}
\Do \Comment{Iterative optimization (quadratic cost approximation)}
	\State $\bm{x}$, $\dot{\bm{x}} \gets \mathrm{rollout} \left( \bm{x}_0, \dot{\bm{x}}_0, \bm{u} \right)$ \Comment{Using \eqref{eq:riem_lqt_vel_update} / \eqref{eq:riem_lqt_pos_update}}
	\State $\bm{L} \bm{L}^\top \gets \bm{\mathcal{S}}'^{\bm{u}\top} \bm{\Gamma}_{\bm{x} \rightarrow \bm{\mu}}^\top \bm{\Sigma}^{-1} \bm{\Gamma}_{\bm{x} \rightarrow \bm{\mu}} \bm{\mathcal{S}}'^{\bm{u}} + \bm{R}$
	\State $\bm{\epsilon} \gets -\mathrm{cholsolve} \left( \bm{\mathcal{S}}'^{\bm{u}\top} \bm{\Gamma}_{\bm{x} \rightarrow \bm{\mu}}^\top \bm{\Sigma}^{-1} \mathrm{Log}_{\hat{\bm{\mu}}}^\mathcal{M} \left( \bm{x} \right) + \bm{R} \bm{u} \right)$
	\State $\bm{u} \gets \bm{u} + \bm{\epsilon}$ \Comment{Update using \eqref{eq:lqt_eps} and line search}
\doWhile{$|| \bm{\epsilon} || > \mathrm{threshold}$}
\State $\bm{V} \gets \bm{L} \backslash \bm{S}^{\bm{u}\top}$ \Comment{Reuse $\bm{L}$ from calculating $\bm{\epsilon}$}
\State $\bm{\Sigma} \gets \bm{V}^\top \bm{V}$ \Comment{Predict covariance for each state \eqref{eq:lqt:cov}}
\end{algorithmic}
\end{algorithm}

\Cref{algo:lqt:fitting} summarizes the steps required for fitting a Riemannian \gls{lqt}.
Note that for computational efficiency, $\bm{\Sigma}_{\bm{u}}$ (required in \eqref{eq:lqt_eps} and \eqref{eq:lqt:cov}) is never explicitly calculated.
Instead, as commonly done in \glspl{gp} \cite{williams2006gaussian,murphy2012machine}, a Cholesky factorization is used to solve the systems of equations thanks to the fact that a covariance matrix is symmetric and positive definite.
Note that in the Euclidean case, convergence is given in one step: evaluating \eqref{eq:lqt_eps} for the first time results in the Euclidean solution \eqref{eq:lqt_eucl_solution} as $\bm{u}$ is zero initially.
With $\mathrm{Log}_{\hat{\bm{\mu}}}^\mathcal{M} \left( \bm{x} \right) = \bm{x} - \hat{\bm{\mu}}$ and $\bm{\Gamma} = \bm{I}$, the Taylor approximation \eqref{eq:lqt_riemann:costfcn_taylor} is exact and $\bm{\epsilon} = \bm{0}$ in the next iteration.

\section[Virtual Fixtures based on Riemannian Linear Quadratic Tracking]{\Acrlongpl{vf} based on Riemannian \gls{lqt}}
\subsection{Virtual Fixture Generation}
\label{sec:method:fixture}
\begin{figure}
	\centering
	\includegraphics[width=\columnwidth,page=4,trim={0 5 0 75},clip]{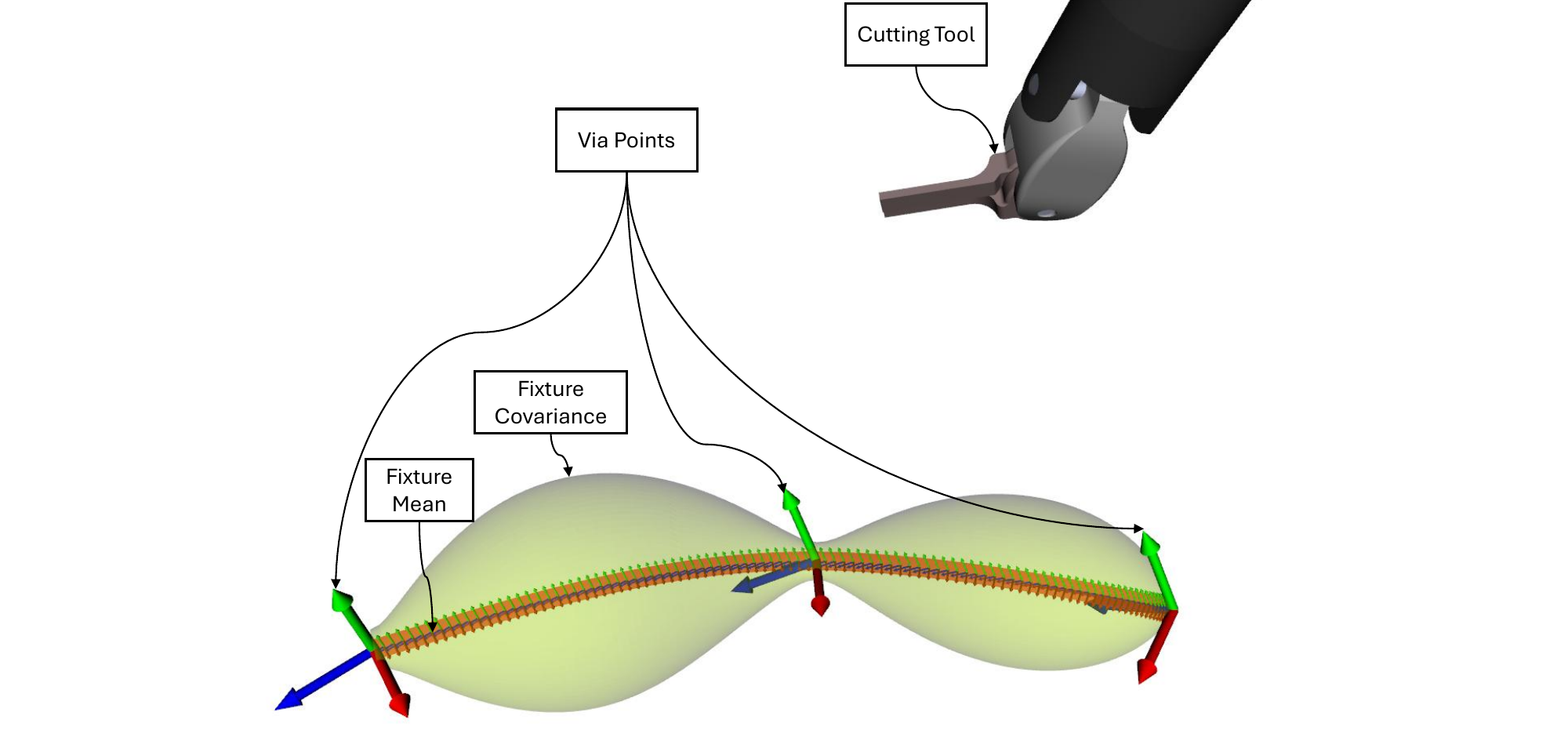}
	\caption{\label{fig:method:lqt_fixture} \Acrlong{vf} generated using the Riemannian \gls{lqt} (\Cref{sec:method:lqt}) from $3$ via points depicted by longer coordinate systems.
	The \gls{lqt} interpolates both position and orientation (shorter coordinate axes) in between; green ellipses show the intersection of the covariance ellipsoid with the plane normal to the trajectory direction at each predicted pose.\vspace{-1em}}
\end{figure}
\Cref{fig:method:lqt_fixture} shows a \gls{vf} created using the \gls{lqt} approach presented in the previous section.
We use the teleoperation system described in the next section without active fixtures to move the remote robot to locations through which the fixture should pass and record the corresponding poses.
Equipped with a predefined covariance and time points linearly interpolated in the interval $[0, 1]$, these poses serve as a reference.
To interpolate a dense trajectory, we calculate the system for $100$ time steps in this interval.
At time points without reference, the only cost comes from the control signal -- the choice of associated cost $\bm{R}$ lets the generated fixture pass more closely through the via points or smoothen the path.
Accordingly, an increased uncertainty and thus larger covariance is associated with these points, as reflected by the larger ellipsoids in \Cref{fig:method:lqt_fixture}.
The extracted poses are sent to the real-time controller~\cite{muehlbauer2025unified} for interpolation and variable impedance control.

\begin{figure*}
\vspace{-1em}
\centering
\begin{subfigure}{0.24\textwidth}
  \centering
  \includegraphics[width=\columnwidth,trim={0 0 0 0},clip]{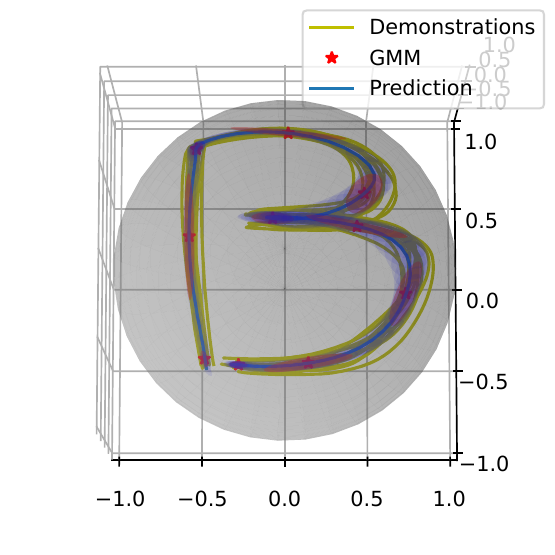}
  \subcaption{\Acrlong{gmr}.\label{fig:eval:plots_s2:gmr}}
\end{subfigure}
\hfill
\begin{subfigure}{0.24\textwidth}
  \centering
  \includegraphics[width=\columnwidth,trim={0 0 0 0},clip]{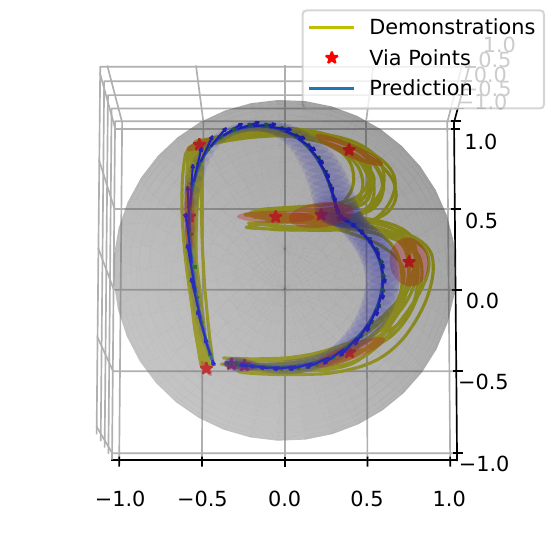}
  \subcaption{LQT with $\bm{R} = \mathbf{I} \cdot 1e-4$.\label{fig:eval:plots_s2:coarse}}
\end{subfigure}
\hfill
\begin{subfigure}{0.24\textwidth}
  \centering
  \includegraphics[width=\columnwidth,trim={0 0 0 0},clip]{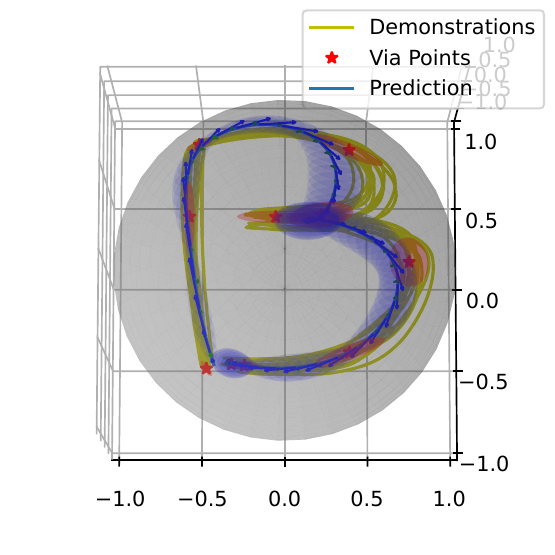}
  \subcaption{LQT with $\bm{R} = \mathbf{I} \cdot 2e-5$.\label{fig:eval:plots_s2:medium}}
\end{subfigure}
\hfill
\begin{subfigure}{0.24\textwidth}
  \centering
  \includegraphics[width=\columnwidth,trim={0 0 0 0},clip]{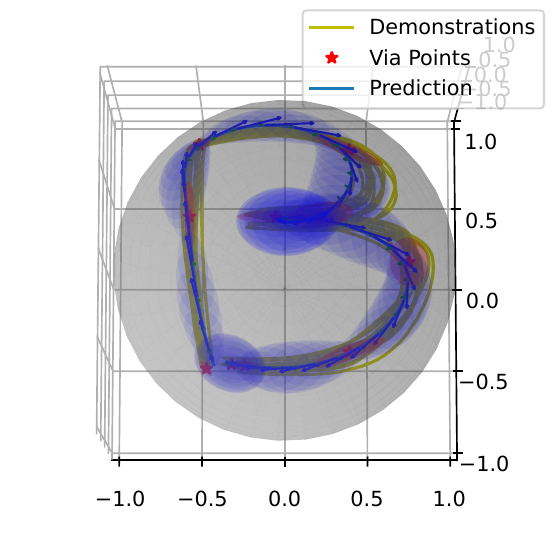}
  \subcaption{LQT with $\bm{R} = \mathbf{I} \cdot 5e-6$.\label{fig:eval:plots_s2:fine}}
\end{subfigure}%
\caption{Plots for writing the letter ``B'' of the handwriting dataset \cite{silverio2023nonparametric} on $\mathcal{S}^2$. We fit the \gls{gmm} using $10$ Gaussians and extract $10$ via points using \gls{gmr} for \gls{lqt}; all these Gaussians are depicted with their mean and covariance ellipsoid. For the \gls{lqt}, we furthermore plot its velocity as blue and the command at each time step as green arrows. Note how it passes through the edge of the covariance ellipsoids only for high control cost, trading accuracy for control cost, while approximating the original trajectory more precisely with decreasing control cost.\vspace{-1em}}
\label{fig:eval:plots_s2}
\end{figure*}

\subsection{Teleoperation System}
\label{sec:method:teleop}
Two modes are important for haptic teleoperation: \textit{coupled} and \textit{clutched} operation.
During the second operation mode, the remote side is kept stationary while the input device can be repositioned freely to achieve an ergonomic position and avoid workspace limits.
When starting the \textit{coupled} mode, the current orientation and position of both devices are stored as $\bm{R}_{\mathrm{offs},\textit{mtm}}$, $\bm{R}_{\mathrm{offs},\textit{psm}}$ and $\bm{t}_{\mathrm{offs},\textit{mtm}}$, $\bm{t}_{\mathrm{offs},\textit{psm}}$, respectively.
Additionally, a relative rotation of the base of the remote side in input coordinates $\bm{R}_{\mathrm{mtm},\mathrm{psm}}$ is required to restore hand-eye coordination by aligning motion directions of \glspl{mtm} and instruments in the display.
Overall, scaled target translation and rotation of the remote \gls{psm} are determined to
\begin{align}
	\bm{R}_{\mathrm{ee},\mathrm{psm}} &= \bm{R}_{\mathrm{offs},\textit{psm}} \Delta \bm{R}_{\mathrm{ee},\mathrm{mtm}}\label{eq:teleop:rotation}\\
	\bm{t}_{\mathrm{ee},\mathrm{psm}} &= \alpha \bm{R}_{\mathrm{mtm},\mathrm{psm}} \Delta \bm{t}_{\mathrm{ee},\mathrm{mtm}} + \bm{t}_{\mathrm{offs},\mathrm{psm}}\label{eq:teleop:translation}
\end{align}
where $\alpha$ is the scaling factor for translations.
Delta values for both orientation and position encode the motion since both sides have been \textit{coupled}, i.e. $\Delta\bm{R}_{\mathrm{ee},\mathrm{mtm}} = \bm{R}_{\mathrm{offs},\mathrm{mtm}}^{-1} \bm{R}_{\mathrm{ee},\mathrm{mtm}}$ and $\Delta\bm{t}_{\mathrm{ee},\mathrm{mtm}} = \bm{t}_{\mathrm{ee},\mathrm{mtm}} - \bm{t}_{\mathrm{offs},\mathrm{mtm}}$.

We use the teleoperated coupling provided by the \gls{dvrk}~\cite{kazanzides2014davinci}, which controls the \glspl{psm} in position mode and the \glspl{mtm} in gravity compensation with wrench overlay, which we employ to add \gls{vf} wrenches.
Note that the \gls{dvrk} always ensures an orientation matching, i.e. setting $\bm{R}_{\mathrm{offs},\mathrm{mtm}} = \bm{I}_3$, $\bm{R}_{\mathrm{offs},\mathrm{psm}} = \bm{R}_{\mathrm{mtm},\mathrm{psm}}$.
To be able to compute \glspl{vf} in \gls{psm} coordinates while applying them to the \gls{psm}, we calculate \eqref{eq:teleop:rotation} and \eqref{eq:teleop:translation} and additionally the remote velocity through \cite{murray2017mathematical}
\begin{align}
	\bm{\omega}_{\mathrm{ee},\mathrm{psm}} &= \bm{R}_{\mathrm{ee},\textit{psm}}^{-1} \bm{R}_{\mathrm{ee},\mathrm{mtm}} \bm{\omega}_{\mathrm{ee},\mathrm{mtm}}\\
	\bm{v}_{\mathrm{ee},\mathrm{psm}} &= \alpha \bm{R}_{\mathrm{ee},\textit{psm}}^{-1} \bm{R}_{\mathrm{ee},\mathrm{mtm}} \bm{v}_{\mathrm{ee},\mathrm{mtm}}.
\end{align}
The wrench for the \gls{mtm} is then rotated inversely \cite{murray2017mathematical}, i.e.
\begin{align}
	\bm{\tau}_{\mathrm{ee},\mathrm{mtm}} &= \bm{R}_{\mathrm{ee},\textit{mtm}}^{-1} \bm{R}_{\mathrm{ee},\mathrm{psm}} \bm{\tau}_{\mathrm{ee},\mathrm{psm}}\\
	\bm{f}_{\mathrm{ee},\mathrm{mtm}} &= \alpha \bm{R}_{\mathrm{ee},\textit{mtm}}^{-1} \bm{R}_{\mathrm{ee},\mathrm{psm}} \bm{f}_{\mathrm{ee},\mathrm{psm}}
\end{align}
and are finally applied by the \gls{dvrk} through $\bm{\tau} = \bm{J}^\top \bm{w}_{\mathrm{ee},\mathrm{mtm}}$ \cite{murray2017mathematical} where $\bm{w}_{\mathrm{ee},\mathrm{mtm}} = \left[ \bm{f}_{\mathrm{ee},\mathrm{mtm}}; \bm{\tau}_{\mathrm{ee},\mathrm{mtm}}\right]$.

\subsection[Virtual Fixture Control]{\Acrlong{vf} Control}
\label{sec:method:control}
With those transformations, we use the trajectory and variable impedance control of \cite{muehlbauer2025unified}, which allows the operator to freely move along the trajectory, and passivate the system using \cite{muehlbauer2026stabilizingarxiv}.
Note the similarity to the adaptive gains of \cite{lembono2021probabilistic} -- \cite{muehlbauer2025unified} additionally guarantees realizable stiffness matrices while the passivation \cite{muehlbauer2026stabilizingarxiv} ensures a smooth entering of fixtures.

For escaping a fixture, we extend the distance-based covariance scaling of \cite{muehlbauer2025unified} using a trigonometric scaling inspired by \cite{selvaggio2018passive} where $d$ is the position distance to the trajectory
\begin{align}
	s =
	\begin{cases}
		1, & d < d_\mathrm{min}\\
		\frac{1}{2} \left( 1 + \mathrm{cos} \left( \pi \frac{d - d_\mathrm{min}}{d_\mathrm{max} - d_\mathrm{min}} \right) \right), & d_\mathrm{min} \le d \le d_\mathrm{max}\\
		0. & d > d_\mathrm{max}
	\end{cases}\raisetag{1.3em}\label{eq:fading_function}
\end{align}
The precision of the current attractor point is scaled with $s$, which in turn scales the stiffness gains.
While this scaling law is already an improvement over the linear scaling found in \cite{muehlbauer2025unified}, deactivating a fixture can still cause fast motions as the operator might not be able to adapt quickly enough for decreasing forces.
We delay decreasing damping matrices calculated through \cite{albuschaeffer2003cartesianimpedance}, which ensures surplus damping that counteracts robot velocities when leaving a fixture and thus smoothens the motion.


\section{Evaluation}
\label{sec:evaluation}

\subsection{Evaluation of the Riemannian LQT}
\label{sec:method:lqt_eval}
In \Cref{fig:eval:plots_s2}, we show our Riemannian \gls{lqt} formulation on the unit sphere $\mathcal{S}^2$, creating trajectories with $100$ points for via points extracted using \gls{gmm} / \gls{gmr} from the handwriting dataset \cite{silverio2023nonparametric}.
Using a large control cost, the coarse trajectory is fitted in \SI{0.02}{\second} with $9$ iterations in a C++ implementation (\Cref{fig:eval:plots_s2:coarse}).
For the medium (\Cref{fig:eval:plots_s2:medium}) and small control cost (\Cref{fig:eval:plots_s2:fine}), we can fit a trajectory in \SI{0.04}{\second} / $10$ iterations and \SI{0.08}{\second} / $11$ iterations, respectively.
Note how the resulting covariance, especially for higher control costs, reflects the covariance of the via points when passing through them while predicting a higher covariance when further away.
This is exactly the behavior we would like to leverage when utilizing \gls{lqt} for modeling \glspl{vf}: at taught poses, we assume \textit{precise} knowledge of the target pose and thus \textit{stiff} guidance, while the higher uncertainty in between is used for \textit{softer} guidance.

\begin{figure}
\centering
\begin{subfigure}{0.24\textwidth}
  \centering
  \includegraphics[width=\columnwidth,trim={0 0 0 0},clip]{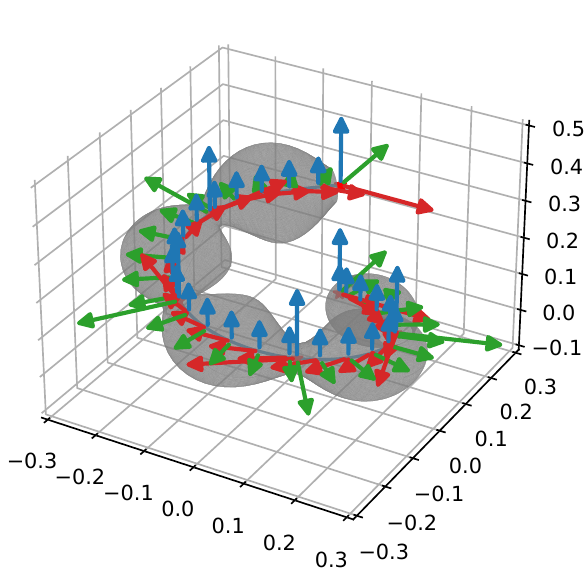}
  \subcaption{Riemannian \gls{lqt}.\label{fig:eval:plots_traj:lqt}}
\end{subfigure}
\hfill
\begin{subfigure}{0.24\textwidth}
  \centering
  \annotategraphicsmulti{
	  \includegraphics[width=\columnwidth,trim={0 0 0 0},clip]{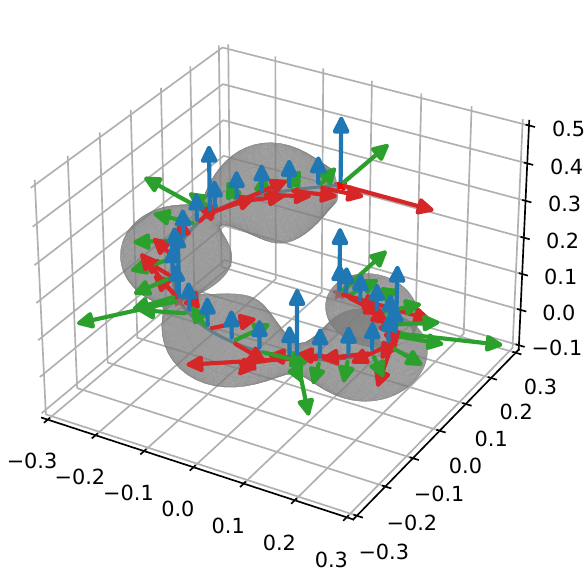}
	}{
        		\node at (0.28,0.38) {F};
    }
  \subcaption{\Acrshort{sts} \gls{lqt} \cite{calinon2020gaussians}.\label{fig:eval:plots_traj:sts_lqt}}
\end{subfigure}%
\caption{Plots of a circular trajectory for our proposed Riemannian \gls{lqt} and a \gls{sts} variant \cite{calinon2020gaussians}. Long coordinate systems show via points and their orientation, short coordinate systems those of the prediction. The gray volume shows the predicted covariance of the position part.}
\label{fig:eval:plots_traj}
\end{figure}

\Cref{fig:eval:plots_traj} shows the predictions of both our proposed Riemannian \gls{lqt} and a \gls{sts} implementation for a circular trajectory including a full orientation change of \SI{360}{\degree}.
While the results on the position part are identical, the \gls{sts} variant at \lref{F} cannot handle orientations at \SI{180}{\degree}, owing to the jump from $-\pi$ to $+\pi$ of the tangent space vector at that orientation.
For modeling full rotations, our formulation is thus indispensable.

\subsection{Pilot Study using the Riemannian LQT Fixture}
\begin{figure}
	\centering
	\includegraphics[width=\columnwidth,page=3,trim={0 300 0 0},clip]{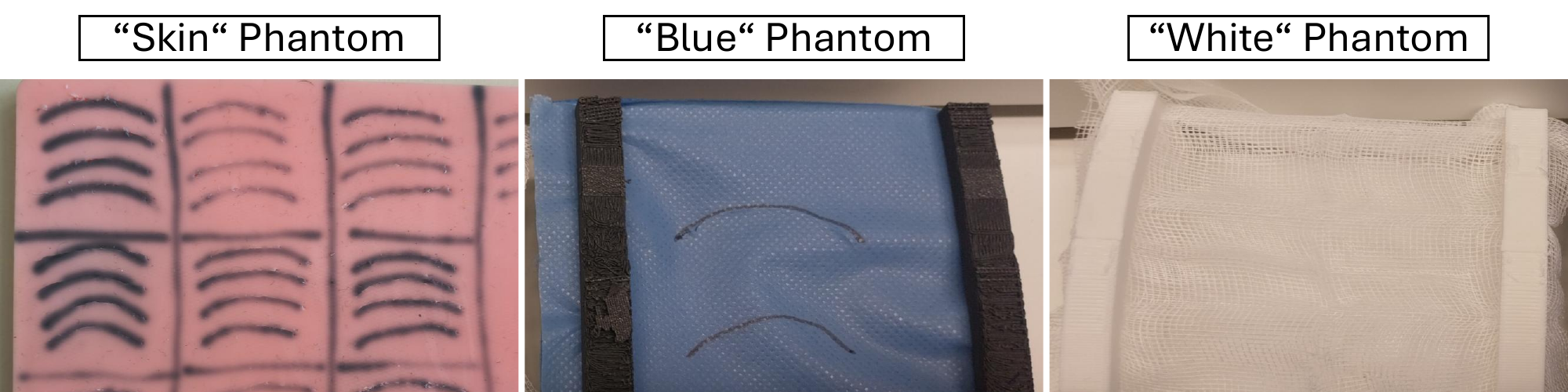}
	\caption{\label{fig:eval:phantoms} Phantoms used in the experiments. The ``skin'' phantom models human skin tissue, the ``blue'' phantom consists of two-layered surgical drape and the ``white'' phantom is multi-layered gauze pad. Of the latter two, only the uppermost layer should be cut.\vspace{-1.5em}}
\end{figure}
We evaluate the effectiveness of our approach in a surgical setting on the task of cutting several phantoms (\Cref{fig:eval:phantoms}) in the setup shown in \Cref{fig:intro:overview}.
For the \textit{skin} and \textit{blue} phantoms, only the right \gls{psm} is required to cut the tissue.
For the \textit{white} phantom, we additionally equip the left \gls{psm} with a gripper which allows for putting tension on the tissue to facilitate cutting.
Leveraging a cylindrical geometry $\mathcal{M}_2$ \cite{muehlbauer2025unified}, we apply an assistive force of \SI{1}{\newton} along the positive radius direction for the manifold centered at the right \gls{psm} to the left \gls{mtm}, thus supporting the user with both collision avoidance and tensioning of the tissue.
At the right \gls{mtm}, we provide force feedback with the method described in the previous sections.

For the teleoperated coupling, we set $\alpha = 0.4$ for the translations in \eqref{eq:teleop:translation}.
In the fading function \eqref{eq:fading_function}, we set $d_\mathrm{min} = 1.5$ and $d_\mathrm{max} = 4$ where $d = (\Delta \bm{x}_{\mathrm{pos}}^\top \bm{\Sigma}_{\mathrm{pos}}^{-1} \Delta \bm{x}_{\mathrm{pos}})^{\frac{1}{2}}$ denotes the unitless Mahalanobis position distance.
To ensure smooth deactivation of the fixture, while realizing the stiffness decrease immediately, we delay the damping decrease by \SI{0.2}{\second}, resulting in an overdamped system compensating for the loss of force counteracting the operator.
Nominal stiffness values are set to $\bm{K}_{p,\mathrm{pos}} = \mathrm{diag}(1000, 1000, 1000) \unit{\newton\per\metre}$ and $\bm{K}_{p,\mathrm{rot}} = \mathrm{diag} (0.2, 0.2, 0.05) \unit{\newton\metre\per\radian}$; with $\lambda^{-} = 0$ and $\lambda^{+} = 10^5$ in the variable stiffness formulation \cite{muehlbauer2025unified}, this stiffness is scaled to half its value between via points for our phantoms.
We determine starting positions of both \glspl{mtm} and \glspl{psm} such that during the whole cutting task, clutching, i.e., a repositioning of the \glspl{mtm}, can be avoided.

For rendering \gls{vf} forces to the user through the \gls{mtm}, we use the force overlay interface provided by the \gls{dvrk}.
The middleware ``links and nodes'' \cite{schmidt2026ln} is used both for process management and communication between the \gls{vf} components which are computed at \SI{200}{\hertz}.
Custom bridges translate communication to ROS 2 used by the \gls{dvrk}.

\subsubsection{Experimental Design}
Our pilot study is conducted with $10$ participants\footnote{During the experiments, the da Vinci system was operated according to its intended use; it was additionally secured through an enable pedal and an emergency cut off, so no special ethical permission is required for the experiments conducted.} ($6$ female, $4$ male; $8$ are medical students with no prior teleoperation experience and $2$ lab members with prior teleoperation experience) aged $23$-$33$ (\SI{25.8 \pm 2.8}{\year}).
The medical students were trained in $2$ preparatory sessions cutting the ``skin'' phantom for a familiarization with the system without the aid of \glspl{vf} prior to the actual experiments.

\begin{figure}
	\centering
	\includegraphics[width=\columnwidth,trim=15 150 20 95,clip]{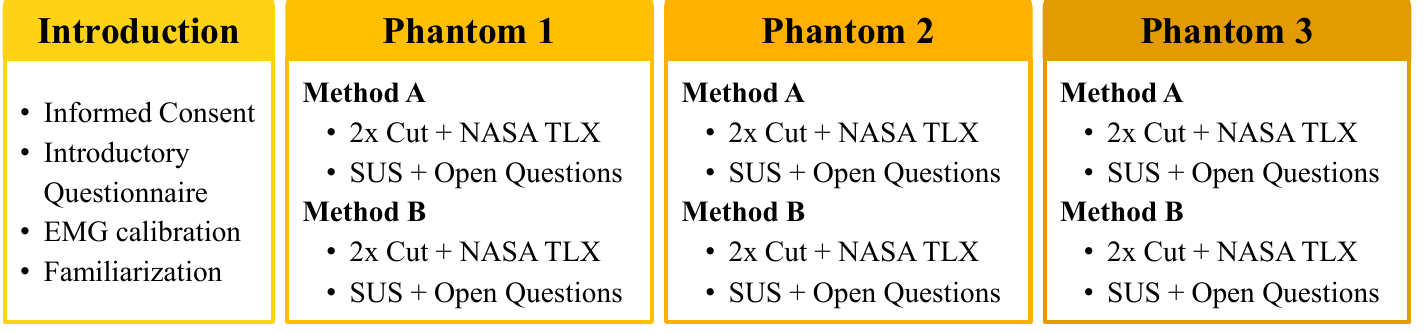}
	\caption{\label{fig:eval:procedure} Experimental procedure for evaluating the proposed method. The phantom order is varied using a latin square design, while half of the participants start \textit{with} and the other half \textit{without} \gls{vf} for \textit{Method A}, \textit{Method B} is the remaining variant.\vspace{-1em}}
\end{figure}
\Cref{fig:eval:procedure} shows the full experimental procedure performed with all participants of the study.
After obtaining \textit{informed consent} and filling an \textit{introductory questionnaire} about demographic information and prior experience, for \textit{\gls{emg} calibration} \cite{lehman1999importance}, the \gls{mvc} of extensor and flexor of both forearms of the subjects are recorded.
Subsequently, in the \textit{familiarization} phase, subjects are shown how to create a \gls{vf} by demonstrating via points.
Then they teach such fixture themselves and use it to perform a trial cut on the ``skin'' phantom.
This procedure is repeated until the participants feel sufficiently prepared for creating and training the fixtures -- for most users, one trial was sufficient whereas some needed two or three trials.

On the test phantoms, the first $5$ participants started \textit{with} \gls{vf} for \textit{Method A} and \textit{without} fixture for \textit{Method B} while the second $5$ participants had this order reversed.
A latin square design was used to permute the order of the three phantoms.
After each trial, subjective workload was measured using the NASA TLX~\cite{hart1988development} questionnaire.
After each method, usability was rated using the SUS~\cite{brooke1996sus} score as well as open questions about the method.
The study was concluded with open questions about the system and potentials for improvement.

\subsubsection{Subjective Results}
\begin{table}
\scriptsize
\centering
\caption{\label{tab:evaluation:sus}\centering \scshape SUS value (M $\pm$ SD) for different phantoms. Higher = better, highlighted in bold.}
\begin{tabular}{@{}c|cccc@{}}
 & Mean & ``Skin'' & ``Blue'' & ``White''\\
\hline
\textit{No Fixture} & $65.0 \pm 13.2$ & $66.8 \pm 7.3$ & $65.0 \pm 17.9$ & $\bf 63.3 \pm 13.4$\\
\textit{Fixture} & $\bf 65.3 \pm 17.4$ & $\bf 67.0 \pm 18.4$ & $\bf 67.0 \pm 18.3$ & $62.0 \pm 17.1$
\end{tabular}
\end{table}

\begin{table}
\scriptsize
\centering
\caption{\label{tab:evaluation:tlx}\centering \scshape Raw NASA TLX (M $\pm$ SD) for different phantoms. First row without, second with fixture, lower = better.}
\begin{tabular}{@{}c|cccc@{}}
 & Mean & ``Skin'' & ``Blue'' & ``White''\\
\hline
No \textit{Fixture} & $6.27 \pm 3.90$ & $5.92 \pm 4.20$ & $\bf 6.00 \pm 3.67$ & $6.88 \pm 3.95$\\
\textit{Fixture} & $\bf 6.01 \pm 2.99$ & $\bf 4.99 \pm 2.55$ & $6.34 \pm 3.03$ & $\bf 6.70 \pm 3.21$\\
\end{tabular}
\vspace{-1em}
\end{table}

Subjective results include the SUS score ($0-100$, higher = better) as shown in \Cref{tab:evaluation:sus} and the NASA TLX score ($0-20$, lower = better) depicted in \Cref{tab:evaluation:tlx}.
When \glspl{vf} are applied, results are in general better, though no significant differences could be found for our sample size of $10$ participants using a repeated measures ANOVA.

In the semi-structured questions on what subjects liked and disliked about the method and what suggestions for improvement they had, results were evenly split on the force feedback of the \gls{vf}.
While one part of the users described the forces experienced as too high and wanted to have more motion freedom, the other part reported the guidance level as close to optimal.
Similarly, users were also split on their level of familiarity with the system - reporting on both ends of good usability of the robot and needing more training.

\subsubsection{Objective Results}
\begin{table}
\scriptsize
\centering
\caption{\label{tab:evaluation:time}\centering \scshape Completion time in seconds for different phantoms. Lower = better, highlighted in bold.}
\begin{tabular}{@{}c|cccc@{}}
 & Mean & ``Skin'' & ``Blue'' & ``White''\\
\hline
\textit{No Fixture} & $\bf 43.2 \pm 26.3$ & $\bf 36.4 \pm 19.7$ & $32.3 \pm 18.9$ & $\bf 60.9 \pm 29.9$\\
\textit{Fixture} & $43.9 \pm 24.5$ & $37.2 \pm 19.9$ & $\bf 28.0 \pm 14.2$ & $66.6 \pm 20.3$
\end{tabular}
\end{table}
\begin{table}
\caption{\label{tab:evaluation:anova_time}\centering \scshape ANOVA comparing completion time results.}
\begin{tabular}{cc|cccc}
Dependent Variable & Factor & df1 & df2 & F & p (corr.)\\
\hline
Completion Time & Fixture & $1$ & $9$ & $0.0684$ & n.s.\\
 & Phantom & $2$ & $18$ & $11.8193$ & $< 0.001$\\
 & Fixt. * Phant. & $2$ & $18$ & $1.2908$ & n.s.
\end{tabular}
\end{table}

We first obtain the objective measure of \textit{completion time} measured from the time point when users first touch the tissue by the tool until the cut is complete (\Cref{tab:evaluation:time}).%
\footnote{Not taking the repetition into account as factor, we averaged over both trials per setting and user. For one user and setting, we could only use the result of one trial as the recording of the other trial failed.}
An ANOVA using the factors ``fixture'' and ``phantom'' as well as their combination (\Cref{tab:evaluation:anova_time}) reveals significant differences for different phantoms, but not for the use of the fixture.
A post-hoc comparison using pairwise t-tests reveals significant differences ($p < 0.001$) between the pairings ``white'' - ``blue'' and ``white'' - ``skin''; the comparison ``blue'' - ``skin'' only shows a non-significant trend ($p < 0.1$).

\begin{table}
\scriptsize
\centering
\caption{\label{tab:evaluation:emg}\centering \scshape \gls{emg} values ($\%$ of \gls{mvc}) for different experiments.}
\begin{tabular}{@{}c|cccc@{}}
 & Mean & ``Skin'' & ``Blue'' & ``White''\\
\hline
Flexor & $\bf 9.9 \pm 19.2$ & $16.3 \pm 31.5$ & $5.9 \pm 3.8$ & $\bf 7.4 \pm 10.8$\\
~~w. \textit{Fixture} & $10.3 \pm 17.8$ & $\bf 15.2 \pm 25.1$ & $\bf 5.6 \pm 3.7$ & $10.2 \pm 18.0$\\
\hline
Extensor & $\bf 9.4 \pm 4.0$ & $\bf 10.5 \pm 3.6$ & $\bf 9.4 \pm 4.5$ & $\bf 8.2 \pm 3.9$\\
~~w. \textit{Fixture} & $10.2 \pm 4.7$ & $11.6 \pm 5.2$ & $9.7 \pm 4.5$ & $9.4 \pm 4.6$\\
\end{tabular}
\end{table}
\begin{table}
\caption{\label{tab:evaluation:anova_emg}\centering \scshape ANOVA results comparing \gls{emg} values.}
\begin{tabular}{cc|cccc}
Dependent Variable & Factor & df1 & df2 & F & p (corr.)\\
\hline
Flexor $\%$ \gls{mvc} & Fixture & $1$ & $9$ & $1.7881$ & n.s.\\
 & Phantom & $2$ & $18$ & $1.6192$ & n.s.\\
 & Fixt. * Phant. & $2$ & $18$ & $0.8406$ & n.s.\\
\hline
Extensor $\%$ \gls{mvc} & Fixture & $1$ & $9$ & $6.7434$ & $<.05$\\
 & Phantom & $2$ & $18$ & $3.9399$ & $<.1$\\
 & Fixt. * Phant. & $2$ & $18$ & $0.8200$ & n.s.
\end{tabular}
\vspace{-1em}
\end{table}

Right forearm \gls{emg}\footnote{Of the $120$ scheduled trials in the analysed dataset, three task exports were absent and four contained no samples. These seven unavailable recordings affected five participants. When one repetition was unavailable, the remaining repetition was used as the participant-level condition estimate. All $10$ participants retained at least one recording in each of the six conditions.} was analysed offline in MATLAB R2024b over the selected cutting intervals.
Signals were filtered using a fourth-order Butterworth band-pass filter with single-pass design frequencies of \SIrange{20}{450}{\hertz}, applied forward and backward for zero-phase filtering.
Whenever a matching full recording was available, filtering preceded extraction of the analysis interval.
A moving RMS envelope was computed using complete \SI{250}{\milli\second} windows.
Mean task RMS was normalized to the maximum RMS of the corresponding MVC recording, processed identically, and expressed as \%MVC.
Valid repetitions were averaged within each participant, phantom and fixture condition.
Separate two-way repeated-measures ANOVAs were performed for the flexor and extensor, with fixture and phantom as within-participant factors.
All main analyses included $10$ participants.
Greenhouse–Geisser corrections were applied to the phantom and interaction effects.
Processing variants and participant exclusions were examined in exploratory sensitivity analyses.
Descriptive statistics and ANOVA results are reported in \Cref{tab:evaluation:emg,tab:evaluation:anova_emg}.
Extensor activation increased from \SI{9.35}{\percent} MVC without the fixture to \SI{10.22}{\percent} MVC with the fixture, with a mean paired difference of $0.87$ percentage points MVC (\SI{95}{\percent} CI $0.11–1.63$; $F(1,9) = 6.7434$, nominal $p < 0.05$).
This increase is expected, as higher forces at the \gls{mtm} are required when correcting a fixture compared to the unconstrained case - the small difference, however, hints that not much correction was required.
No significant fixture effect was detected for the flexor ($F(1,9) = 1.7881$).
The extensor phantom effect did not reach the \SI{5}{\percent} threshold after Greenhouse–Geisser correction, and neither interaction was significant.

\begin{table}
\scriptsize
\centering
\caption{\label{tab:evaluation:dist_center}\centering \scshape Distance between cut and marked line in mm. Lower = better, highlighted in bold.}
\begin{tabular}{@{}c|ccc@{}}
 & Mean & ``Skin'' & ``Blue''\\
\hline
\textit{No Fixture} & $1.1 \pm 0.7$ & $1.1 \pm 0.5$ & $1.1 \pm 0.8$\\
\textit{Fixture} & $\bf 0.9 \pm 0.6$ & $\bf 0.9 \pm 0.5$ & $\bf 0.9 \pm 0.7$\\
\end{tabular}
\end{table}
\begin{table}
\scriptsize
\centering
\caption{\label{tab:evaluation:dist_len}\centering \scshape Distance between start / end of cut and marked line start / end in mm. Lower = better, highlighted in bold.}
\begin{tabular}{@{}c|ccc@{}}
 & Mean & ``Skin'' & ``Blue''\\
\hline
\textit{No Fixture} & $1.1 \pm 0.9$ & $1.1 \pm 0.8$ & $1.1 \pm 1.0$\\
\textit{Fixture} & $\bf 0.7 \pm 0.7$ & $\bf 0.9 \pm 0.6$ & $\bf 0.6 \pm 0.7$\\
\end{tabular}
\vspace{-1em}
\end{table}
\begin{table}
\caption{\label{tab:evaluation:anova_distances}\centering \scshape ANOVA results comparing results with and without fixture, excluding the ``white'' phantom.}
\begin{tabular}{cc|cccc}
Dependent Variable & Factor & df1 & df2 & F & p\\
\hline
Cut Offset & Fixture & $1$ & $9$ & $4.3675$ & $<.1$\\
 & Phantom & $1$ & $9$ & $0.0377$ & n.s.\\
 & Fixt. * Phant. & $1$ & $9$ & $0.0149$ & n.s.\\
\hline
Cut Length & Fixture & $1$ & $9$ & $6.7924$ & $<.05$\\
 & Phantom & $1$ & $9$ & $0.5015$ & n.s.\\
 & Fixt. * Phant. & $1$ & $9$ & $0.8161$ & n.s.\\
\end{tabular}
\end{table}

Next, we compare two distance metrics: the \textit{distance between cut and marked line} (\Cref{tab:evaluation:dist_center}), measured as maximum deviation perpendicular of the cut line, and the \textit{distance between start / end of cut and marked line} (\Cref{tab:evaluation:dist_len}), measured as the distance the actual cut is shorter or longer than the marked line.
Due to the nature of the phantoms, we can only accurately measure these distance measures for the ``blue'' and ``skin'' phantoms, we thus omit this measure for the ``white'' phantom.
Distances when using the fixture are lower for both phantoms, an ANOVA (\Cref{tab:evaluation:anova_distances}) reveals a non-significant trend ($p < 0.1$) for the \textit{distance between cut and marked line} (denoted as ``Cut Offset'') and a significant difference ($p < 0.05$) for the \textit{distance between start / end of cut and marked line} (denoted as ``Cut Length'').

\subsubsection{Discussion}
In the pilot study, we compared our proposed approach using \glspl{vf} defined by the user from few via points with a baseline using pure teleoperation.
Even though already the pure teleoperation showed a high usability in the subjective metrics, we found a significant difference for the cut length and a non-significant trend for the cut offset.
With both metrics serving as proxy to task accuracy, we can conclude that our method allows even non-expert users to create guides that are actually useful for task success.

Overall, we believe our approach could benefit from more training of the users to aid them in creating more useful guidance; furthermore, this would allow us to adapt parameters such as stiffness values to the individual user preferences.
To this end, tests with advanced surgeons could reveal the usefulness in real clinical scenarios and allow for a more fine-grained evaluation of subtle properties of our method such as the variable stiffness formulation.

\section{Conclusion and Outlook}
\label{sec:conclusion}
We have presented an approach for the creation of probabilistic \Acrlongpl{vf} from only few demonstrated via points.
To enable a geometry-aware guidance including orientation, we have extended \Acrlong{lqt} to a geometry-aware formulation on Riemannian manifolds.
Thanks to computational improvements using the Cholesky decomposition, we achieve fast runtimes of the algorithm despite requiring iterations for fitting.
In the overall teleoperation system, we leverage covariance predictions for varying the stiffness of the impedance controller and calculate the control algorithms in coordinates of the remote robot, transforming the resulting wrenches to the local side to provide user feedback.

A pilot study involving $10$ subjects shows that the method improves task performance regarding accuracy metrics, thus providing real benefits even for inexperienced users.
Future work should involve adaptation of the method parameters to user preferences and a more extensive evaluation.
Vision could also be used to extract control points for the creation of guidance, leveraging detection uncertainty to control the covariance of individual via points.
Additionally, we see potential benefits in surgical training scenarios where experts create guides that allow novice surgeons to faster learn specific surgical skills.

\section*{Acknowledgments}
\label{sec:acknowledgements}
This work was partially funded by the DLR project ``ASPIRO''; the European Union’s Horizon Research and Innovation Program under Grant 101136067 (INVERSE) and the BRIEF ``Biorobotics Research and Innovation Engineering Facilities'' project (Project identification code IR0000036).

\enlargethispage{\baselineskip}

\bibliographystyle{./style/IEEEtran}
\bibliography{literatur}

\begin{thebibliography}{10}
\providecommand{\url}[1]{#1}
\csname url@samestyle\endcsname
\providecommand{\newblock}{\relax}
\providecommand{\bibinfo}[2]{#2}
\providecommand{\BIBentrySTDinterwordspacing}{\spaceskip=0pt\relax}
\providecommand{\BIBentryALTinterwordstretchfactor}{4}
\providecommand{\BIBentryALTinterwordspacing}{\spaceskip=\fontdimen2\font plus
\BIBentryALTinterwordstretchfactor\fontdimen3\font minus
  \fontdimen4\font\relax}
\providecommand{\BIBforeignlanguage}[2]{{%
\expandafter\ifx\csname l@#1\endcsname\relax
\typeout{** WARNING: IEEEtran.bst: No hyphenation pattern has been}%
\typeout{** loaded for the language `#1'. Using the pattern for}%
\typeout{** the default language instead.}%
\else
\language=\csname l@#1\endcsname
\fi
#2}}
\providecommand{\BIBdecl}{\relax}
\BIBdecl

\bibitem{rosenberg1993virtualfixtures}
L.~Rosenberg, ``Virtual fixtures: Perceptual tools for telerobotic
  manipulation,'' in \emph{Proceedings of IEEE Virtual Reality Annual
  International Symposium}, 1993, pp. 76--82.

\bibitem{bowyer2014active}
S.~A. Bowyer, B.~L. Davies, and F.~R. y~Baena, ``Active constraints/virtual
  fixtures: A survey,'' \emph{{IEEE} Transactions on Robotics}, vol.~30, no.~1,
  pp. 138--157, Feb 2014.

\bibitem{muehlbauer2022multiphase}
M.~Mühlbauer, F.~Steinmetz, F.~Stulp, T.~Hulin, and A.~Albu-Schäffer,
  ``Multi-phase multi-modal haptic teleoperation,'' in \emph{2022 IEEE/RSJ
  International Conference on Intelligent Robots and Systems (IROS)}.\hskip 1em
  plus 0.5em minus 0.4em\relax IEEE, 10 2022.

\bibitem{raiola2017comanipulation}
G.~Raiola, S.~S. Restrepo, P.~Chevalier, P.~Rodriguez-Ayerbe, X.~Lamy,
  S.~Tliba, and F.~Stulp, ``Co-manipulation with a library of virtual guiding
  fixtures,'' \emph{Autonomous Robots}, vol.~42, no.~5, pp. 1037--1051, 11
  2017.

\bibitem{zeestraten2018programming}
M.~J.~A. Zeestraten, I.~Havoutis, and S.~Calinon, ``Programming by
  demonstration for shared control with an application in teleoperation,''
  \emph{IEEE Robotics and Automation Letters}, vol.~3, no.~3, pp. 1848--1855,
  2018.

\bibitem{muehlbauer2024probabilistic}
M.~Mühlbauer, T.~Hulin, B.~Weber, S.~Calinon, F.~Stulp, A.~Albu-Schäffer, and
  J.~Silvério, ``A probabilistic approach to multi-modal adaptive virtual
  fixtures,'' \emph{IEEE Robotics and Automation Letters}, vol.~9, no.~6, pp.
  5298--5305, 2024.

\bibitem{kazanzides2014davinci}
P.~Kazanzides, Z.~Chen, A.~Deguet, G.~S. Fischer, R.~H. Taylor, and S.~P.
  DiMaio, ``An open-source research kit for the da vinci® surgical system,''
  in \emph{2014 IEEE International Conference on Robotics and Automation
  (ICRA)}.\hskip 1em plus 0.5em minus 0.4em\relax IEEE, 2014.

\bibitem{zeestraten2017manifold}
M.~J.~A. Zeestraten, I.~Havoutis, J.~Silvério, S.~Calinon, and D.~G. Caldwell,
  ``An approach for imitation learning on {R}iemannian manifolds,'' \emph{IEEE
  Robotics and Automation Letters}, vol.~2, no.~3, pp. 1240--1247, 2017.

\bibitem{calinon2020gaussians}
S.~Calinon, ``Gaussians on riemannian manifolds: Applications for robot
  learning and adaptive control,'' \emph{IEEE Robotics \& Automation Magazine},
  vol.~27, no.~2, pp. 33--45, 2020.

\bibitem{pruks2022method}
V.~Pruks and J.-H. Ryu, ``Method for generating real-time interactive virtual
  fixture for shared teleoperation in unknown environments,'' \emph{The
  International Journal of Robotics Research}, vol.~41, no. 9–10, pp.
  925--951, 2022.

\bibitem{selvaggio2016enhancing}
M.~Selvaggio, G.~Notomista, F.~Chen, B.~Gao, F.~Trapani, and D.~Caldwell,
  ``Enhancing bilateral teleoperation using camera-based online virtual
  fixtures generation,'' in \emph{2016 IEEE/RSJ International Conference on
  Intelligent Robots and Systems (IROS)}.\hskip 1em plus 0.5em minus
  0.4em\relax IEEE, 10 2016.

\bibitem{menoth2025review}
D.~Menoth~Mohan, B.~Shirinzadeh, J.~Smith, Y.~Zhong, S.~H. Turlapati, and
  D.~Campolo, ``A review of recent methods and applications of virtual fixtures
  in robot-assisted surgery,'' vol. 199, p. 111294.

\bibitem{selvaggio2018passive}
M.~Selvaggio, G.~A. Fontanelli, F.~Ficuciello, L.~Villani, and B.~Siciliano,
  ``Passive virtual fixtures adaptation in minimally invasive robotic
  surgery,'' \emph{IEEE Robotics and Automation Letters}, vol.~3, no.~4, pp.
  3129--3136, 2018.

\bibitem{moccia2019vision}
R.~Moccia, M.~Selvaggio, L.~Villani, B.~Siciliano, and F.~Ficuciello,
  ``Vision-based virtual fixtures generation for robotic-assisted polyp
  dissection procedures,'' in \emph{2019 IEEE/RSJ International Conference on
  Intelligent Robots and Systems (IROS)}.\hskip 1em plus 0.5em minus
  0.4em\relax IEEE, 2019, pp. 7934--7939.

\bibitem{calinon2015tutorial}
S.~Calinon, ``A tutorial on task-parameterized movement learning and
  retrieval,'' \emph{Intelligent Service Robotics}, vol.~9, no.~1, pp. 1--29,
  Sep 2015.

\bibitem{rozo2020learning}
L.~Rozo, M.~Guo, A.~G. Kupcsik, M.~Todescato, P.~Schillinger, M.~Giftthaler,
  M.~Ochs, M.~Spies, N.~Waniek, P.~Kesper, and M.~Burger, ``Learning and
  sequencing of object-centric manipulation skills for industrial tasks,'' in
  \emph{2020 IEEE/RSJ International Conference on Intelligent Robots and
  Systems (IROS)}.\hskip 1em plus 0.5em minus 0.4em\relax IEEE, 2020, pp.
  9072--9079.

\bibitem{ferro2023coppelia}
M.~Ferro, A.~Mirante, F.~Ficuciello, and M.~Vendittelli, ``A coppeliasim
  dynamic simulator for the da vinci research kit,'' \emph{IEEE Robotics and
  Automation Letters}, vol.~8, no.~1, pp. 129--136, 2023.

\bibitem{muehlbauer2025unified}
\BIBentryALTinterwordspacing
M.~Mühlbauer, B.~Weber, S.~Calinon, F.~Stulp, A.~Albu-Schäffer, and
  J.~Silvério, ``A unified framework for probabilistic dynamic-, trajectory-
  and vision-based virtual fixtures,'' 2025. [Online]. Available:
  \url{https://arxiv.org/abs/2506.10239}
\BIBentrySTDinterwordspacing

\bibitem{fletcher2012geodesic}
P.~Thomas~Fletcher, ``Geodesic regression and the theory of least squares on
  riemannian manifolds,'' \emph{International Journal of Computer Vision}, vol.
  105, no.~2, pp. 171--185, 2012.

\bibitem{kim2014multvariate}
H.~J. Kim, N.~Adluru, M.~D. Collins, M.~K. Chung, B.~B. Bendin, S.~C. Johnson,
  R.~J. Davidson, and V.~Singh, ``Multivariate general linear models (mglm) on
  riemannian manifolds with applications to statistical analysis of diffusion
  weighted images,'' in \emph{2014 IEEE Conference on Computer Vision and
  Pattern Recognition}.\hskip 1em plus 0.5em minus 0.4em\relax IEEE, 2014, pp.
  2705--2712.

\bibitem{ti2023geometric}
B.~Ti, A.~Razmjoo, Y.~Gao, J.~Zhao, and S.~Calinon, ``A geometric optimal
  control approach for imitation and generalization of manipulation skills,''
  \emph{Robotics and Autonomous Systems}, vol. 164, p. 104413, 2023.

\bibitem{lachner2020influence}
J.~Lachner, V.~Schettino, F.~Allmendinger, M.~D. Fiore, F.~Ficuciello,
  B.~Siciliano, and S.~Stramigioli, ``The influence of coordinates in robotic
  manipulability analysis,'' \emph{Mechanism and Machine Theory}, vol. 146, p.
  103722, 2020.

\bibitem{hogan1984impedance}
N.~Hogan, ``Impedance control: An approach to manipulation,'' in \emph{1984
  American Control Conference}.\hskip 1em plus 0.5em minus 0.4em\relax IEEE, 07
  1984.

\bibitem{muehlbauer2026stabilizingarxiv}
\BIBentryALTinterwordspacing
M.~M{\"u}hlbauer, N.~Werner, R.~Balachandran, T.~Hulin, J.~Silv{\'e}rio,
  F.~Stulp, and A.~Albu-Sch{\"a}ffer, ``Passive variable impedance for shared
  control,'' \emph{arXiv preprint arXiv:2604.20557}, 2026. [Online]. Available:
  \url{https://arxiv.org/abs/2604.20557}
\BIBentrySTDinterwordspacing

\bibitem{albuschaeffer2003cartesianimpedance}
A.~Albu-Schäffer, C.~Ott, U.~Frese, and G.~Hirzinger, ``Cartesian impedance
  control of redundant robots: recent results with the dlr-light-weight-arms,''
  in \emph{2003 IEEE International Conference on Robotics and Automation},
  vol.~3, 2003, pp. 3704--3709 vol.3.

\bibitem{lembono2021probabilistic}
T.~S. Lembono and S.~Calinon, ``Probabilistic iterative lqr for short time
  horizon mpc,'' in \emph{2021 IEEE/RSJ International Conference on Intelligent
  Robots and Systems (IROS)}.\hskip 1em plus 0.5em minus 0.4em\relax IEEE, pp.
  579--585.

\bibitem{williams2006gaussian}
C.~K. Williams and C.~E. Rasmussen, \emph{Gaussian processes for machine
  learning}.\hskip 1em plus 0.5em minus 0.4em\relax MIT press Cambridge, MA,
  2006, vol.~2.

\bibitem{murphy2012machine}
K.~P. Murphy, \emph{Machine learning: a probabilistic perspective}.\hskip 1em
  plus 0.5em minus 0.4em\relax MIT press, 2012.

\bibitem{silverio2023nonparametric}
J.~Silvério and Y.~Huang, ``A non-parametric skill representation with soft
  null space projectors for fast generalization,'' in \emph{2023 IEEE
  International Conference on Robotics and Automation (ICRA)}, 2023, pp.
  2988--2994.

\bibitem{murray2017mathematical}
R.~M. Murray, Z.~Li, and S.~S. Sastry, \emph{A Mathematical Introduction to
  Robotic Manipulation}.\hskip 1em plus 0.5em minus 0.4em\relax CRC Press,
  2017.

\bibitem{schmidt2026ln}
F.~Schmidt, J.~Nix, M.~Mühlbauer, J.~Cremer, M.~Chalon, T.~Bachmann, and
  A.~Raffin, ``Links and nodes: Middleware for distributed real-time robotic
  systems,'' \emph{Journal of Open Source Software}, vol.~11, no. 124, p.
  10777, 2026.

\bibitem{lehman1999importance}
G.~J. Lehman and S.~M. McGill, ``The importance of normalization in the
  interpretation of surface electromyography: A proof of principle,''
  \emph{Journal of Manipulative and Physiological Therapeutics}, vol.~22,
  no.~7, pp. 444--446, 1999.

\bibitem{hart1988development}
S.~G. Hart and L.~E. Staveland, ``Development of nasa-tlx (task load index):
  Results of empirical and theoretical research,'' in \emph{Advances in
  psychology}.\hskip 1em plus 0.5em minus 0.4em\relax Elsevier, 1988, vol.~52,
  pp. 139--183.

\bibitem{brooke1996sus}
J.~Brooke \emph{et~al.}, ``Sus-a quick and dirty usability scale,''
  \emph{Usability evaluation in industry}, vol. 189, no. 194, pp. 4--7, 1996.

\end{thebibliography}
\end{document}